\documentclass[conference]{IEEEtran}
\IEEEoverridecommandlockouts
\usepackage{cite}
\usepackage{amsmath,amssymb,amsfonts}
\usepackage{algorithmic}
\usepackage{graphicx}
\usepackage{textcomp}
\usepackage{xcolor}
\usepackage{url}
\usepackage{tablefootnote}
\usepackage{todonotes}
\usepackage{float}

\def\BibTeX{{\rm B\kern-.05em{\sc i\kern-.025em b}\kern-.08em
    T\kern-.1667em\lower.7ex\hbox{E}\kern-.125emX}}
\begin{document}

\title{Leveraging 5G private networks, UAVs and robots to detect and combat broad-leaved dock (Rumex obtusifolius) in feed production}

\author{\IEEEauthorblockN{Christian Schellenberger\IEEEauthorrefmark{1}, Christopher Hobelsberger\IEEEauthorrefmark{1}, Bastian Kolb-Grunder\IEEEauthorrefmark{1}, Florian Herrmann\IEEEauthorrefmark{2} \\
		and Hans D. Schotten\IEEEauthorrefmark{1}\IEEEauthorrefmark{2}}
\IEEEauthorblockA{
\textit{\IEEEauthorrefmark{1}Institute for Wireless Communication and Navigation} \\
\textit{Rheinland-Pf\"alzische Technische Universit\"at Kaiserslautern-Landau}\\
Kaiserslautern, Germany \\
E-Mail: \{christian.schellenberger, b.kolbgrunder, c.hobelsberger, schotten\}@rptu.de\\ \\
\IEEEauthorblockA{
\textit{\IEEEauthorrefmark{2}Intelligent Networks} \\
\textit{Deutsches Forschungszentrum f\"ur K\"unstliche Intelligenz}\\
Kaiserslautern, Germany \\
E-Mail: \{florian.herrmann, schotten\}@dkfi.de}
}}

\maketitle

\begin{abstract}
In this paper an autonomous system to detect and combat Rumex obtusifolius leveraging autonomous unmanned aerial vehicles (UAV), small autonomous sprayer robots and 5G SA connectivity is presented. Rumex obtusifolius is a plant found on grassland that drains nutrients from surrounding plants and has lower nutritive value than the surrounding grass. High concentrations of it have to be combated in order to use the grass as feed for livestock. One or more UAV are controlled through 5G to survey the current working area and send back high-definition photos of the ground to an edge cloud server. There an AI algorithm using neural networks detects the Rumex obtusifolius and calculates its position using the UAVs position data. When plants are detected an optimal path is calculated and sent via 5G to the sprayer robot to get to them in minimal time. It will then move to the position of the broad-leafed dock and use an on-board camera and the edge cloud to verify the position of the plant and precisely spray crop protection only where the target plant is. The spraying robot and UAV are already operational, the training of the detection algorithm is still ongoing. The described system is being tested with a fixed private 5G SA network and a nomadic 5G SA network as public cellular networks are not performant enough in regards to low latency and upload bandwidth.
\end{abstract}


\section{Introduction}
Rumex obtusifolius is a perennial weed that competes with pasture species in grassland. It is estimated that in Germany seven out of 10 grassland farms are considered to have serious problems in 1992 with Rumex obtusifolius or Rumex crispus \cite{ZAL04}. It not only displaces valuable forage grasses but also affects the quality of basic forage and its preservation possibilities. It can be eaten by grazing animals, but it has less nutritive value than surrounding grasses \cite{OSW83}. The seeds of the plant can survive in the soil for up to 50 years and thus represent a long-term problem \cite{HER09}.
\subsection{Combating Rumex obtusifolius}
The two main methods for controlling Rumex on agricultural land are mechanical and chemical control of the plants. Where mechanical control is always selective and chemical control is mostly used on the whole target field.

In mechanical control, i.e., weeding, the plants are individually picked out from the subsoil. This procedure is very labor intensive and takes about 280 h/ha \cite{LfL22}. Therefore, it is only profitable when infestation is low. In recent years automated weeding robots are being developed. The development is largely driven by organic farming practices, where the use of synthetic herbicides is not allowed \cite{BEL21}. The downside of automated weeding robots is their mechanical complexity. Due to this maintenance costs can be higher than for spraying robots.

Chemical control is mostly done with selective herbicides across the whole area, but also can be selective, i.e., each plant can be treated individually with a paint stick or backpack sprayer. The labor required for individual plant control is comparable to the mechanical method. Large-scale application of herbicides has the disadvantage of high soil contamination by chemicals and soil compaction due to heavy duty tractors. Another downside of chemical control, especially for large-scale application, is the waiting time between the application and the next cut of the grass.% For maximum effectiveness it is important to wait for the perfect climate conditions. There should not be rain right before the application as well as three to four hours after. The ambient temperature should be above 5°C \cite{POE01}. 
\subsection{Vision based detection of rumex obtusifolius}
There are many different methods already described in literature how to detect weeds including rumex obtusifolius in images. In \cite{GEB06} a ground-based contraption was used to take pictures at a height of 1.60 m. These pictures were processed to calculate feature information which was used to classify the them based on a maximum likelihood approach. Segmenting areal pictures into smaller squares and classifying them using AI is another approach discussed in \cite{LAM21}. Pictures were taken with the UAV at different heights with the best results at a height of 10 m. The segmented squares had a size of 0.5 m x 0.5 m and were classified based on them containing rumex obtusifolius or not using a VGG16 neural network \cite{SIM14}. The predictions were post processed using feature information to reduce false positives. \cite{EVE09} uses Fourier analysis to describe the texture of pictures. Grass shows a larger contribution of high-frequency basis functions to the total signal than rumex obtusifolius. This allows them to detect the more homogeneous texture of rumex obtusifolius compared to the inhomogeneous texture of gras in pictures. 
\subsection{Automated weed control}
There are several different entities working on automated combat of weeds in general and Rumex obtusifolius specifically. \cite{EVE11} describes a robot that uses GPS to drive a pre-defined route and detects broad-leaved dock through machine vision. When a plant is detected it is destroyed using a cutting device.  The robot described in \cite{BLA02} also uses machine vision to detect weeds in a vegetable crop field. Once detected the weed is then killed with an electric discharge of 15 kV. The system is pulled by a commercial tractor and due to the high electricity demand is not suited as a standalone system. The system presented in \cite{WU20} shows a way to alleviate the need for real-time detection of weeds by using multiple non-overlapping cameras. The detected weeds are then sprayed with herbicides. Due to the non-overlapping nature of the cameras the robot has to be relatively large. Using computer vision for detection and a sprayer nozzle attached to a manipulator arm \cite{UND15} demonstrates a system quite similar to the one presented in this paper, but doesn't use UAVs to support the detection of the weed therefore taking longer to treat the whole field. To decrease the time needed to treat the whole field they also suggest using a swarm of robots. AgBotII \cite{BAW17} is an integrated robot with machine vision to discern different weeds. It uses different chemical or mechanical methods for the weed destruction depending on the weed species. It is larger than the system described in this paper and also doesn't use UAVs. The use of smaller machinery in agriculture follows a broader trend in agriculture to use smaller equipment as seen in project Xaver by Fendt \cite{FEN20}. They are more flexible when fields vary greatly in size, cost less and don't compact the soil compared to heavy tractors. 

There are also commercial robots that have the goal of controlling docks, for example, there is EcoRobotix \cite{ECO22} with its autonomous robot AVO. The battery-powered robot combats weeds with herbicides and has a maximum area output of 10 hectares per day. The robot is battery-powered and its operating time is extended by solar cells. Its weight of 750 kg minimizes soil compaction. Wi-Fi and 4G are also integrated, but these interfaces are used exclusively for monitoring and manual intervention. Tensorfield Agriculture \cite{TEN22} is another provider of a dock combating system, but takes a different approach. Their robot uses a method called "thermal micro dosing". Here, very hot vegetable oil is sprayed on weeds to destroy the plants. However, both the image recognition and the control of the robot are done entirely on the robot. The machine can cover 10 hectares with one tank of oil.
\subsection{5G in agriculture}
5G will enable new applications in agriculture. Since 5G networks are not yet widespread the body of work concerning 5G an agriculture is not very big yet. In \cite{TAN21} a broad vision of future applications enabled by 5G are discussed.\cite{RAZ19} shows a UAV based system using 5G to offload the data to a central video analytics server to monitor crops and livestock. Their 5G network is using TV White Space (TVWS) technology to test the potential of shared radio spectrum for 5G in rural areas. With this limitation, upload and download of around 50 -- 60 Mbit/s were achieved. Using this network, the full video resolution was only achieved 11\% of the time. In \cite{VAL19} the possibilities of 5G, mobile edge computing and robotics in agriculture are highlighted and framework based on different entities like UAVs, field sensors etc. is proposed. Two possible use cases, autonomous harvesting robots and UAV monitoring using First-Person-View (FPV) drones are discussed.

5G will enable new applications, especially when fast transmission of large amounts of data and low latency communication are required. Remote control applications are conceivable to control one's fleet of agricultural machines or even to let them work semi- or fully autonomously. Which will become especially important in regards to the demographic change where automation can alleviate to a declining labor pool. According to \cite{SMI20} the European labor pool will decrease by 13.5 million (or 4\%) by 2030. In addition, 5G offers the possibility of outsourcing compute intensive applications, like AI detection, to the edge cloud. The centralization of computing power makes the individual machines less expensive, as they only need to carry the basic sensor and transmission technology and not high computing power which scales much better cost-wise in data centers. This approach also decreases the energy demand and therefore extends the working time of individual machines \cite{RAU21}.

%Massive Machine Types Communication (mMTC) is one of the three central application profiles of 5G NR. It enables ahigh density of low energy IoT devices in one cell. Small sensors can be used to collect soil data (moisture, temperature, mineral content, etc.) or livestock health.

Compared to public cellular networks private 5G networks can be adapted to fit a lot of different applications. Cellular networks have the advantage of predictable network access times through centralized medium access control instead of best effort medium control in Wi-Fi and can therefore achieve lower and more reliable latencies. Using dedicated spectrum, cellular networks do not suffer from interference from surrounding networks and have greater coverage area with one base station. The drawback of cellular networks is higher capex and opex costs compared to enterprise Wi-Fi. Private cellular networks provide high configurability compared to public cellular networks, but require highly trained staff to operate. An overview of the advantages and considerations for private 5G networks can be found in \cite{ANG20}. Table \ref{tab1:CellularPerformance} shows the performance of different cellular networks.

\begin{table*}[h]
	\centering
	\caption{Performance data of cellular networks.}
	\label{tab1:CellularPerformance}
	\begin{tabular}{|p{13em}|p{7.5em}|p{7.5em}|p{7.5em}|}
		\hline
		&  private 5G SA & public 4G \cite{OOK21} & public 5G NSA \\
		\hline
		%Max. theoret. throughput & \textgreater 10 Gbit/s & $\sim$ 1 Gbit/s & $\sim$ 9.6 Gbit/s\\
		Avg. throughput Downlink & $\sim$ 700 Mbit/s $^{\mathrm{1}}$ & $\sim$ 90 Mbit/s & $\sim$ 240 Mbit/s $^\mathrm{2}$\\
		Avg. throughput Uplink & $\sim$ 300 Mbit/s $^{\mathrm{1}}$ & $\sim$ 18 Mbit/s & $\sim$ 110 Mbit/s $^\mathrm{2}$\\
		Latency & $\sim$ 10 ms $^{\mathrm{1}}$ & $\sim$ 25 ms & $\sim$ 20 ms $^\mathrm{2}$\\
		%Distance & $\sim$ 300 m - 10 km & $\sim$ 50 m - 15 km & $\sim$ 50m\\
		\hline
		\multicolumn{4}{l}{$^{\mathrm{1}}$Measured in a private 5G SA Rel. 15 network with 100MHz bandwidth (3.7-3.8GHz) with Upload/Download ratio 3:7 on campus of RPTU} \\
		\multicolumn{4}{l}{$^{\mathrm{2}}$Measured in Kaiserslautern 22.11.2021}
	\end{tabular}
\end{table*}
%The big advantage of a private 5G SA network over public mobile networks or Wi-Fi is that the full bandwidth is in an exclusively allocated frequency range. This enables higher data rates and more reliable transmission. In general, the values in Table \ref{tab1:CellularPerformance} should be viewed with caution, since the data rates achieved depend on many factors, such as the number of antennas, terminal device, release version, etc.

\section{System description}
\subsection{System Architecture}
\begin{figure}[h]
	\includegraphics[width=0.44\textwidth]{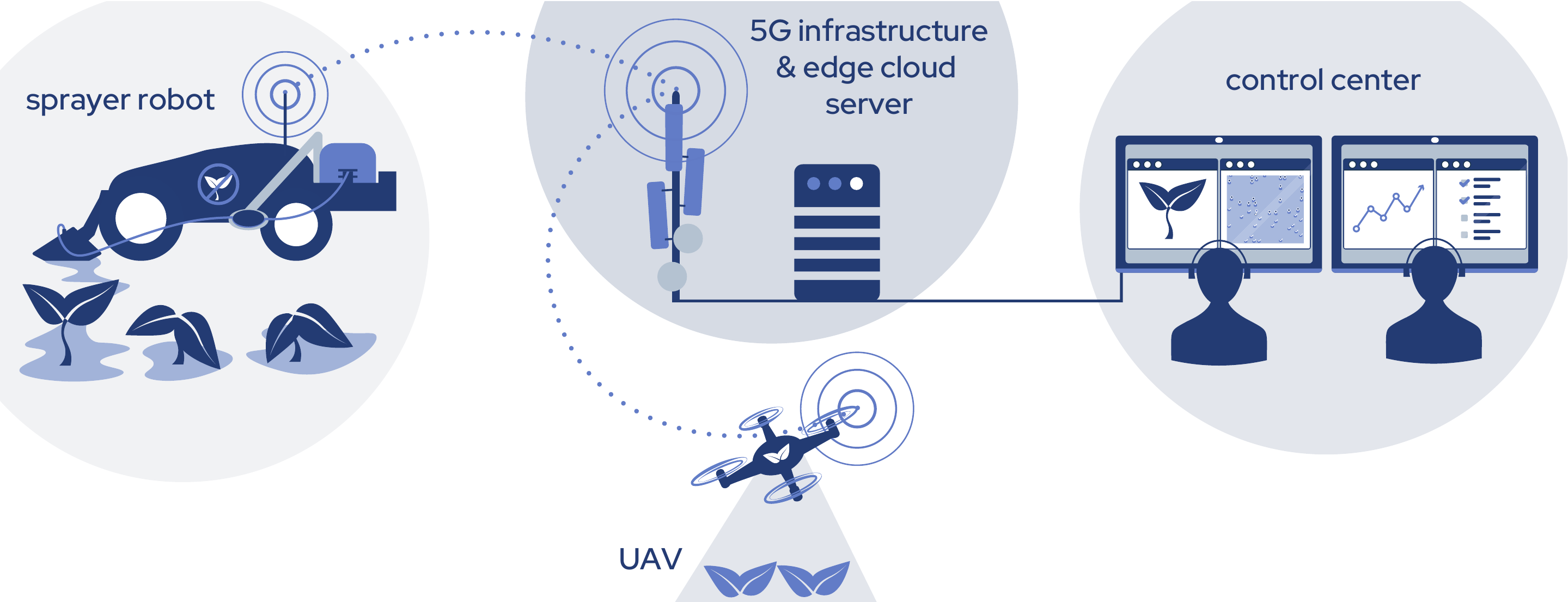}
	\caption{Overall system architecture.} \label{fig:Architecture}
\end{figure}

Figure \ref{fig:Architecture} shows the architecture and communication of the overall system . For efficient detection and precise application of herbicides several sub-systems are needed. They are: UAVs for areal detection, sprayer robots for close-range detection and precise application of herbicide, edge cloud server for centralized high-power computing tasks, a control center for monitoring and controlling all vehicles and systems, and a 5G SA network for handling all communications between edge cloud and the vehicles.

\subsection{UAV platform}
One of the core elements of the use case for detecting rumex on agricultural land agricultural areas is an autonomous drone equipped with a camera and a 5G modem. The drone takes high-resolution images of the agricultural land, which are then sent over a 5G network to the edge cloud for analysis by a high-performance AI cf. \ref{subsec:rumexdet}.% For a later productive use, sufficient flight time for the analysis of large agricultural areas must be ensured. The motorization of the drone must be dimensioned according to the maximum take-off weight including the attachments for rumex detection. 

\begin{figure}[h]
	\includegraphics[width=0.44\textwidth]{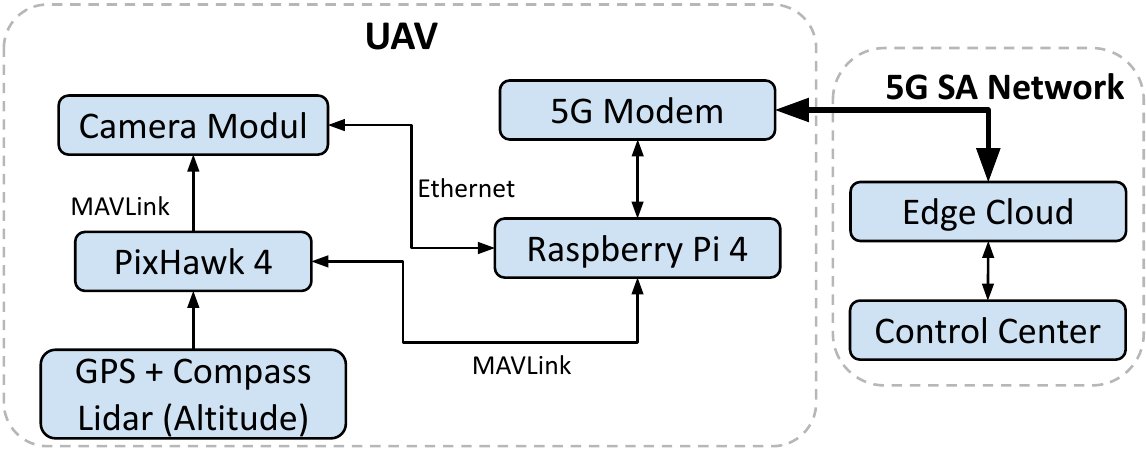}
	\caption{Architecture of the UAV.} \label{fig:detection_uav}
\end{figure}

Currently a Tarot Hexa-Copter frame as the basis of our flying system is used. The overall architecture of the UAV is shown in Figure \ref{fig:detection_uav}. A PixHawk 4 flight controller running the latest Ardupilot Copter firmware is controlling the UAV. It is communicating with all onboard systems and the control software in the edge cloud using the Micro Air Vehicle Link (MAVLink) protocol. The main control and telemetry link is realized through a MAVProxy service which is running on a Raspberry Pi 4 that is connected through a telemetry port on the Pixhawk \cite{KOU19}. A Quectel RM500Q-GL 5G modem connected to the Raspberry Pi allows high speed and low latency communication with the edge cloud. This high-speed link is utilized to establish a MAVLink connection to a client and to transfer pictures and corresponding metadata to the detection AI in the edge cloud. For a broad overview of the flight path, an additional FPV camera was installed. The live video feed is processed by the Raspberry Pi 4 and then transmitted through the 5G network back to the control center. FPV systems are either analog or digital. Analog systems have low latency, but low resolution video feeds. Currently used digital FPV systems have high resolution, but high latency feed. The 5G network enables high resolution and low latency video feeds. Furthermore, it is possible to stream multiple feeds at the same time over the same network connection.   

A radio control link is available at all times as a backup system and by legal requirement for open category UAVs \cite{EU19}. In the current configuration flight time is 15-25 min. The drone gets its position and heading from a Holybro GPS system and a dedicated compass module. The flight height is measured by GPS and a Lidar sensor. The Lidar is also being used to realize a true terrain following. 

High-spatial-resolution areal images, are needed to detect even the smallest rumex plants. A pixel size of 0.05 cm is optimal \cite{ZHA21}\cite{CER21}. The drone flies at a speed of 3 m/s along the preplanned path taking pictures along the way. Speed, heading and altitude are tracked by the Raspberry Pi to calculate the capture time of each individual picture. The pictures have an overlap of 10\% resulting in a width of 3.93 m covered by each track. This results in an average time of about 14 minutes to cover a field of 1 ha (ca. 4.2 ha/h). The detection camera on the drone takes pictures with a resolution of 12.3 Megapixel. These pictures are tagged using GPS and heading data for later evaluation. The precise heading and GPS information is obtained from the flight controller of the drone via a local MAVLink connection. Each picture's midpoint is defined through a GPS coordinate. This information in combination with the logged heading and altitude is later used to calculate the position of each detected rumex cluster. Each picture is currently uncompressed and has a size of approximately 192 Mbit. At an average speed of 3 m/s the drone captures a picture about every 820 ms. This results in a data traffic of about 240 Mbit/s for the detection camera. The drone's FPV camera generates a H.264 video stream with 1080p at 60 fps which results in an additional data traffic of about 6 Mbit/s. In the future compression of the pictures taken by the UAV will be investigated, but due to computational cost has not yet been implemented to increase flight time.

%Path optimization is used to reduce the time and energy consumption it takes to spray all rumex plants detected in the former step. The edge cloud has the required computing power to calculate an optimized sequence in which the sprayer robot approaches the rumex clusters. This sequence of points on which clusters were detected is sent to the robot over the 5G network. We consider a path with minimal length as optimal because the robot would take the least amount of time to follow it. To calculate the shortest path connecting all points is a classic traveling salesman problem which is known to be np hard. We are currently evaluating two possible heuristics to calculate the shortest path the first is a greedy approach taking always the nearest neighbor. The other is the Lin-Kerninghan heuristic. To evaluate the performance of these heuristics we also calculate the optimal path given the number of detected rumex clusters is in range where it is feasible to do so. 

\subsection{Sprayer robot}
Each rumex plant, which is detected on the pictures taken by the drone, has to be treated with a herbicide. For this a sprayer robot is used. In addition to applying herbicides the robot is scanning the ground in front of the sprayer bars to ensure that only rumex plants are sprayed.

The field robot is divided in two main sections, the chassis and the sprayer attachment on top. The chassis houses 4 electric motors, lead-acid batteries, electric speed controllers, main motor controller, wheel encoders and an RC receiver. The attachment on top includes the main computing unit running robot operating System 2 (ROS2) \cite{ROS2} incl. a  Quectel RM500Q-GL 5G modem, Lidar scanners, GPS, inertial measurement unit (IMU), Jetson AGX Xavier, close range cameras, solenoid valve control board, herbicides pump, herbicide tanks, fine mesh inline filter, solenoid valves and nozzles.

To achieve a maximum modularity and portability a collapsible and dividable sprayer bar is used. The main sprayer contains six nozzles and has a working width of 550 mm. On each side one additional sprayer arm with 5 nozzles can be added. The maximum working width is 1800 mm. Each sprayer bar has its own close range detection camera, for the first testing period each camera has its own computing unit on board. Later on each close-range video feed will be sent over the 5G network to the edge cloud to be processed. This will reduce the cost of the robot as well as the power consumption which will lead to a longer run-time. 

\begin{figure}[h]
	\begin{center}
		\includegraphics[width=0.44\textwidth]{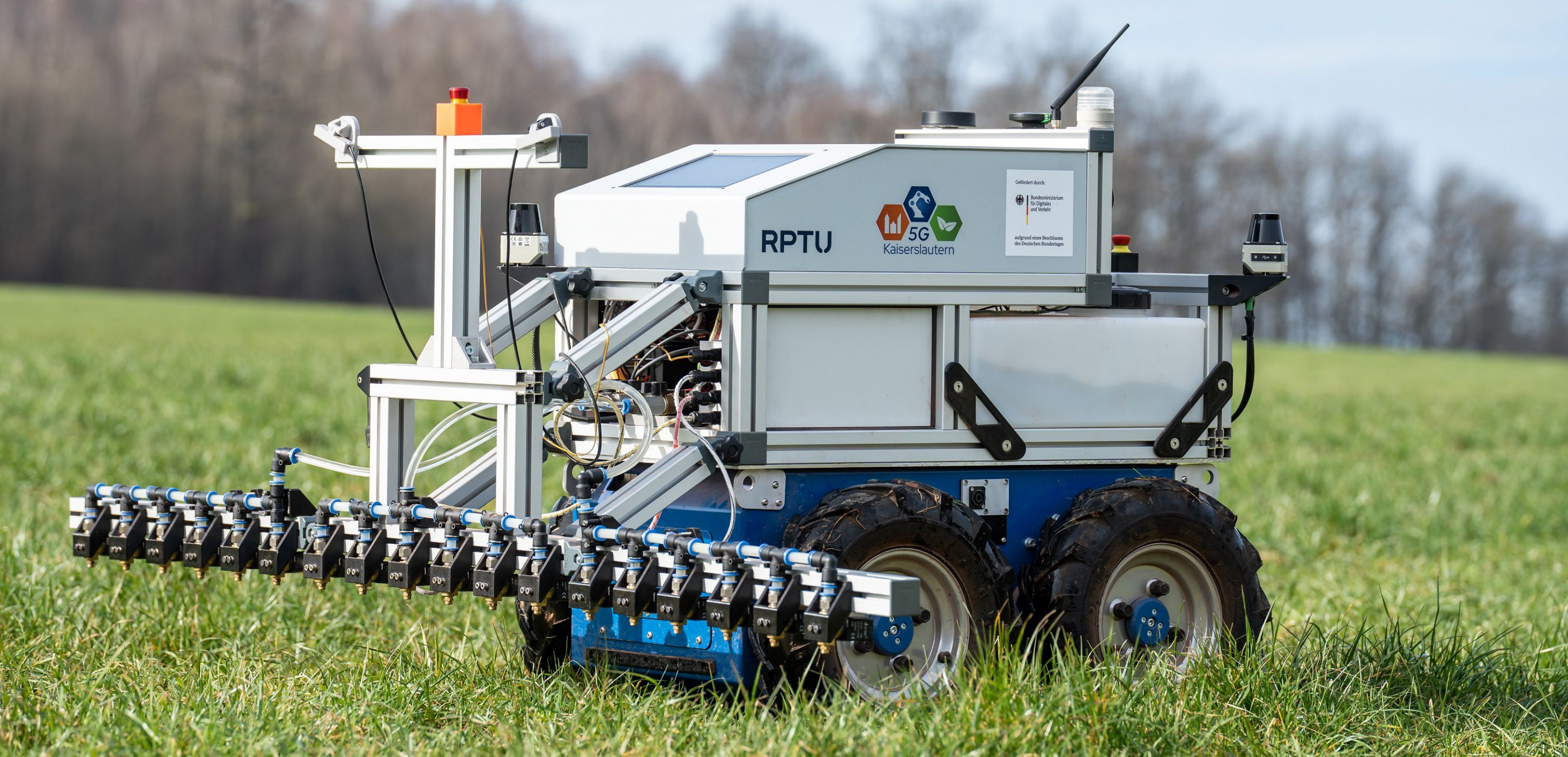}
		\caption{Architecture of the sprayer robot.} \label{fig:sprayer_robot}
	\end{center}
\end{figure}

\subsubsection{Navigation}
An optimized path, calculated in the edge cloud, is transmitted to the field robot through the 5G network to reduce the time and energy it takes to spray all detected rumex plants. We consider a path with minimal length as optimal. To calculate the shortest path connecting all points is a classic traveling salesman problem which is known to be NP-hard. We are currently evaluating two possible heuristics to calculate the shortest path. The first is a greedy approach taking always the nearest neighbor. The other is the Lin-Kerninghan heuristic \cite{LIN73}. To determine its position in the field the robot utilizes fused sensor data coming from its encoders, IMU and GPS. The Nav2 framework \cite{MAC20} integrated into robot OS uses a costmap as an environment representation. With the lidar sensors the robot can detect obstacles and add them to the costmap so the local path planning can avoid them. Between plants the robot moves with a speed of 2 m/s. To improve the quality of detection the speed is reduced to 0.5 m/s when close to the detected plant.
%An optimized path is transmitted to the field robot through the 5G network.  The robot runs ROS2 including the Nav2 navigation framework.  To determine its position in the field the robot utilizes fused sensor data coming from its encoders, IMU and GPS. The Nav2 framework uses a so called costmap as environment representation. This grid-based map is used to plan an optimal local path. With the lidar sensors the robot can detect obstacles and add them to the costmap so the local path planning can avoid them. Between clusters the robot moves with a speed of 2m/s. When the robot arrived at a cluster it covers the area of the cluster in a circular pattern. To improve the quality of detection the speed is reduced to 0,5 - 1m/s during this process. To increase soil protection the navigation framework is configured to avoid turning in place and prefer curved paths to target points. 
\subsubsection{Close range detection}
The coordinates of potential rumex plants are detected by the AI on the edge cloud. Several factors are influencing the precision of the location of the plant detected by the UAV, like GPS and compass accuracy. To detect the rumex plants as precise as possible a close range-video-detection system is installed on the robot. This system is covering the area right in front of the sprayer bar and detects the rumex plants within the working area. For now each camera is connected to a Jetson AGX Xavier on the robot which is running its own AI to detect rumex.
\subsubsection{Precision herbicide application}
To minimize the amount of herbicides needed the sprayer robot uses 16 individual controllable fan nozzles. Each nozzle has a working width of approximately 150 mm and an overlap of 10 mm at a working height of 250 mm above ground. Each nozzle is directly connected to a solenoid valve. This reduces the latency between the time the valve is opened and the time pressure is constant at the tip of the nozzle.

As soon as a rumex plant in front of a nozzle is detected, the correct time to open the corresponding nozzle(s) is being calculated. Each plant is different in size and shape, which influences the time one or more nozzles are activated. Each nozzle opens 20 mm in front of a rumex plant and closes 20 mm after passing over it. This ensures that the whole plant is covered and a minimum amount of herbicides is applied. 

To calculate how many plants can be sprayed with one tank, we estimate the size of an average rumex plant to be about 100 mm in diameter and the average speed of the sprayer robot (in detecting mode) is 0.5 m/s. Each nozzle has to be activated 140 mm of travel which is equal to 0.28 s. At 3 bars the installed nozzles have a throughput of 15 ml/s.  A total of 4.2 ml is needed to spray one plant. With the installed 24 L tank it would be possible to spray a maximum of 5700 plants. 

Each herbicide needs to be diluted to a different concentration to achieve maximum effectiveness. It is important to differentiate between precise punctual application and imprecise large-area covering application. The herbicide concentration must be adjusted so that the previously mentioned 4.2 ml that are sprayed on a plant are sufficient and within the guidelines of the applied herbicide \cite{POE01}. If the concentration of the herbicide is not high enough for 4.2 ml per plant the speed of the robot has to be decreased further.
\subsection{Rumex detection}
\label{subsec:rumexdet}
There are two AIs in use. One AI is used to detect rumex plants in pictures taken by the UAV. The second one is used to detect rumex plants in front of the robot using video streams.

\subsubsection{Areal detection}For the UAV pictures, an AI based on a convolutional neural network was trained with a database of areal pictures of rumex obtusifolius. The AI predicts bounding boxes around the rumex plants using the RGB values of the pictures as input. From these bounding boxes the midpoint is calculated and used for the path planning. To reduce false positives, the bounding boxes need to have a minimum area as well as a certain area to length ratio. Removing bounding boxes with a too small area from the set of detected plants reduces the number of false positives, but it could also remove correctly detected very small plants. As mentioned in \cite{POE01} the rumex needs to have a certain size to be combated effectively so this poses no significant problem. At the moment the system is tested with tennis balls as a representation of plant locations and a pretrained AI for detection as the rumex database is in the process of being created as currently rumex is not growing on the fields. This AI was per-trained using a residual network \cite{HE16} trained with the COCO dataset \cite{LIN14}. 

\subsubsection{Ground detection}As the control of the sprayer robots nozzles are much more time critical than the detection of plants on the UAV pictures, the AI network used to detect plants on pictures form the sprayer robot is different. For this task we use a YOLOv3 \cite{RED18} network because it provides a good mix between performance and accuracy. The neural network works on video streams captured form the detection cameras. It was trained on close up videos of rumex obtusifolius and predicts bounding boxes similar to the AI used for the UAV pictures. As each camera picture is aligned to 6 nozzles the X range of the bounding boxes determines which nozzles are activated and the Y range determines the timing of the activation.
\subsection{Control Center}
The control center allows the user to control and monitor the whole process conveniently from a central point. From here a mission for the UAV can be planned in advance. For path planning we use “Mission Planner”. The path is planned to cover the field in parallel tracks at a height of 10 m. A mission contains a start location, take-off and landing instructions as well as the planned path. Furthermore, it is possible to interact directly with the UAV and the robots through the 5G network.  

The location of the detected rumex plants are displayed on a map in the control center. This allows the user to determine the infestation density of the field. As soon as the UAV finishes the sweep of the field  an optimized path is calculated and sent to the sprayer robot.

The operator is able to see the location of the UAV and the robots at all times on the map. Furthermore, the operator can monitor all sensor data from the robot and observe the planned path in advance. Because 5G offers low latency each vehicle can be stopped immediately by the operator from the control center or automatically through onboard safety controls. After clarifying the cause of the automated emergency stop, the operator can resume the mission. To resolve problems in cases of emergency the operator can take direct control of the UAV and the robot \cite{CER21}.

%Each vehicle has multiple cameras installed. There are two cameras on the UAV, one 1080p FPV camera and the 12.3 MP camera for the detection system. Each picture that is taken by the UAV is uncompressed and has a size of approximately 192 Mbit. At an average speed of 3 m/s the UAV captures a picture about every 880 ms. This results in a data traffic of about 240 Mbit/s for the detection camera. The UAV's FPV camera generates a h264 video stream with 1080p at 60 fps which results in an additional data traffic of about 6 Mbit/s. In addition to the image traffic a small amount of data is generated by the MAVLink telemetry. All data streams are transmitted over the 5G network to the edge cloud server for further processing. The combined upstream traffic of the UAV is at the current upper limit of transmission speed. The sprayer robot has between two to four cameras, depending on the configuration. The setup includes one 1080p FPV camera on the robot and one 1080p camera on each sprayer bar for close-range rumex detection. The close-range detection cameras produce a compressed video stream with a fixed bitrate of 50 Mbit/s.  The FPV camera produces a much more compressed video stream generating a traffic of an additional 6 Mbit/s. With a change to symmetrical upload/download time slots the onboard detection of the sprayer robot can also be offloaded into the edge cloud. Otherwise, the compression of the pictures for the detection has to be compressed. 

\subsection{5G network}
For the presented trials private 5G SA networks are used. The current public network coverage of the farmland in the trial is not sufficient in latency and bandwidth to be used for the described application. Currently two options for the 5G network are available. The first option is a permanently installed antenna close to the grassland area where the trials take place. It is directly connected to the internal network of farm. The edge cloud server for the image processing is located in the small server room in the basement of the farm's main building. The control room has a direct fiber connection to the 5G core router for easy data access. The second option for the 5G SA network is a nomadic network integrated in a light commercial vehicle. The edge server is in the same rack as the cellular network components. The vehicle has a small integrated control station for manually controlling the operation of the drone and robots. It can be powered for a limited time with the integrated batteries or through shore power (fixed installation or with a generator). Depending on the size and distribution of the farmland and the future prices of cellular networking equipment it can be more cost effective to install fixed masts for cellular communication.

\begin{figure}[h]
	\includegraphics[width=0.44\textwidth]{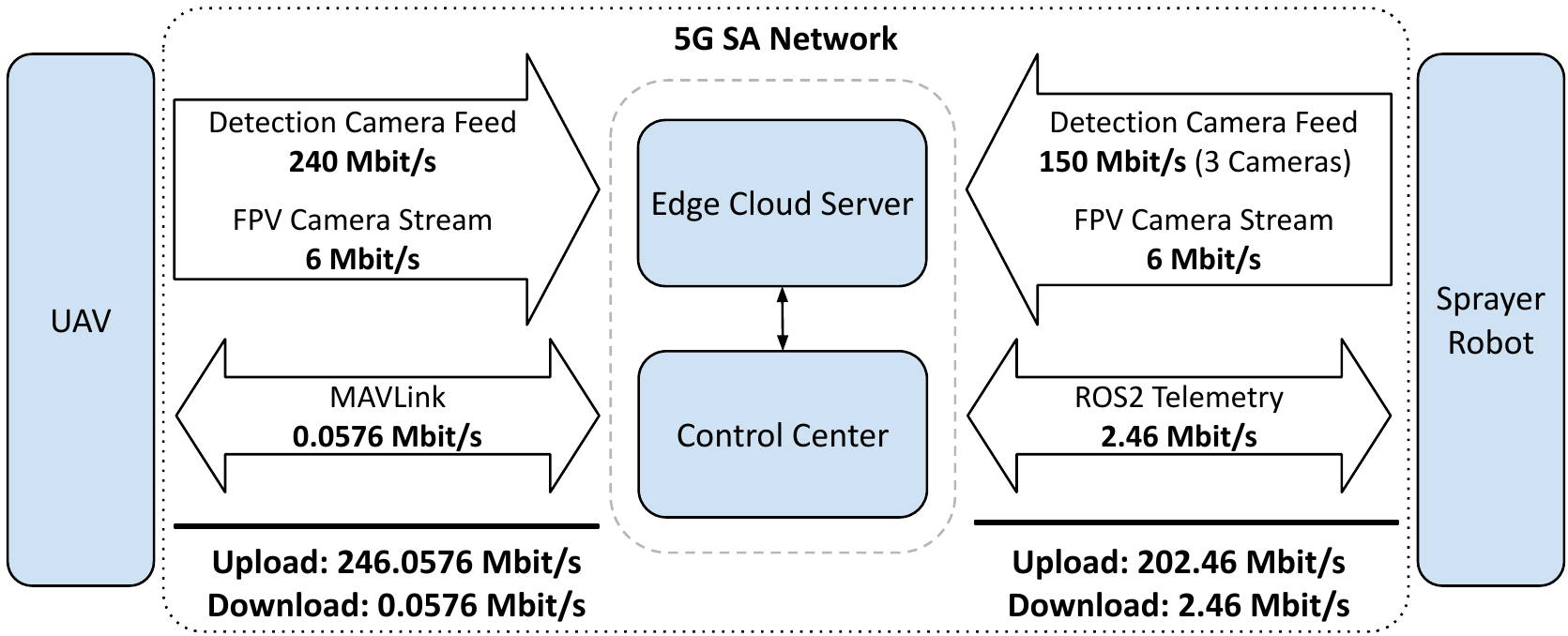}
	\caption{Overall throughput requirements for one UAV and one sprayer robot} \label{fig:DataRates}
\end{figure}

For the network traffic for the overall application is shown in figure \ref{fig:DataRates}. Our current network configuration supports up to 300 Mbit/s upload speeds, which are exceeded by required upload speeds. To solve this problem the data from the UAV can be lossless compressed or pre-processed using the Raspberry Pi on the UAV. The compression of the camera videos from the sprayer robot can also be be increased. With further system updates we expect to be able to achieve 1:1 upload/download speeds which should increase the upload speed to $\sim$ 500 Mbit/s which would allow uncompressed video transmission.

\section{Conclusion}
This paper presents an autonomous system to detect and combat Rumex obtusifolius leveraging private 5G SA networks for areal detection using UAVs, small autonomous sprayer robots to navigate to the detected plants. The robot confirms the location of the target plants and uses precision nozzels to spray the docks. The first test results show that the overall system architecture is feasable and all components work as expected. Private 5G SA networks can be used for the communication between UAV and edge cloud, and edge Cloud to sprayer robot. The raw data of the UAV can be sent to an edge cloud and processed there, which can not be done with Wifi due to range or 4G due to bandwidth limitations. 

Our system and approach to combat rumex obtusifolius selectively in combination with UAV, small autonomous sprayer robots and a 5G SA network will reduce the amount of herbicide needed drastically and therefore provide benefits to nature, ecosystems and farmers.
%The exact localization using on the sprayer robot still currently uses onboard detection, but will be switched to edge cloud in the near future. We expect this data to be more latency sensitive than the UAV data as the robot has to be able to open the sprayer nozzles in time. The 

%The first tests results showed that the current setup is working. The first measurements displayed a great need for a high bandwidth connection. In a next phase all data streams should be transmitted through the 5G SA network to transfer all necessary data from the UAV and sprayer robot to the edge cloud server. To control and monitor the UAV and the sprayer robot through the 5G SA network, not only a high bandwidth is required, a low latency connection is needed as well. Currently the latency of the 5G SA network is not <1ms, this is needed to remotely control the UAV. At this point only mission data is transferred through the 5G SA Network. Future releases for the 5G network will improve low-latency applications.  
\section{Future work}
The exact localization using on the sprayer robot still currently uses onboard detection, but will be switched to edge cloud in the near future. With this energy can be saved to extend operating time and reduce the cost per vehicle. We expect this data to be more latency sensitive than the UAV data as the robot has to be able to open the sprayer nozzles in time. The latency requirements have to be analyzed and feasibility tested with the available network. Further innovations in 5G Rel. 16 will be analyzed for their use in the presented application. Other methods than using herbicides to combat the rumex could also be explored.

%In the future, the project will collect both resilient network-related performance data and analyze and evaluate agricultural results for longer periods. Performance indicators such as availability and data rate can then be compared with previous mobile network generations. Plans are also underway to scale the robot and UAV fleet so that area performance can be increased. In particular, the innovation 5G Sidelink, which will be implemented with Rel. 16, will open up new possibilities for applications in agriculture. Direct communication between the individual platforms can thus be implemented without detours via the mobile network. 
\section{Acknowledgments}
The authors acknowledge the financial support by the Federal Ministry of Digital and Transport of Germany in the program “5x5G-Strategie” (project number VB5GFKAISE). The authors thank CSEM and Agroscope for providing sample data sets of Rumex obtusifolius for AI training.

\bibliographystyle{IEEEtran}
\bibliography{Literature}

\begin{thebibliography}{10}
\providecommand{\url}[1]{#1}
\csname url@samestyle\endcsname
\providecommand{\newblock}{\relax}
\providecommand{\bibinfo}[2]{#2}
\providecommand{\BIBentrySTDinterwordspacing}{\spaceskip=0pt\relax}
\providecommand{\BIBentryALTinterwordstretchfactor}{4}
\providecommand{\BIBentryALTinterwordspacing}{\spaceskip=\fontdimen2\font plus
\BIBentryALTinterwordstretchfactor\fontdimen3\font minus
  \fontdimen4\font\relax}
\providecommand{\BIBforeignlanguage}[2]{{%
\expandafter\ifx\csname l@#1\endcsname\relax
\typeout{** WARNING: IEEEtran.bst: No hyphenation pattern has been}%
\typeout{** loaded for the language `#1'. Using the pattern for}%
\typeout{** the default language instead.}%
\else
\language=\csname l@#1\endcsname
\fi
#2}}
\providecommand{\BIBdecl}{\relax}
\BIBdecl

\bibitem{ZAL04}
J.~Zaller, ``Ecology and non-chemical control of rumex crispus and r.
  obtusifolius (polygonaceae): a review,'' \emph{Weed research}, vol.~44,
  no.~6, pp. 414--432, 2004.

\bibitem{OSW83}
A.~Oswald and R.~Haggar, ``The effects of rumex obtusifolius on the seasonal
  yield of two mainly perennial ryegrass swards,'' \emph{Grass and Forage
  Science}, vol.~38, no.~3, pp. 187--191, 1983.

\bibitem{HER09}
M.~Hermle, A.~Schaller, H.~Thalmann, and H.~Dierauer,
  ``Ampferregulierung-vorbeugende m{\"o}glichkeiten aussch{\"o}pfen,'' 2009.

\bibitem{LfL22}
\BIBentryALTinterwordspacing
\relax Bayerische Landesanstalt~f{\"u}r Landwirtschaft. Ampferregulierung im
  {E}inzelpflanzenbehandlungsverfahren ({V}ersuchsprogramm 934). [Online].
  Available:
  \url{https://www.lfl.bayern.de/mam/cms07/ips/dateien/versuchsbericht_934-03.pdf}
\BIBentrySTDinterwordspacing

\bibitem{BEL21}
\relax Bundesministerium f{\"u}r Ern{\"a}hrung~und Landwirtschaft~(BMEL).
  (2021) {\"O}kologischer {L}andbau in {D}eutschland.

\bibitem{GEB06}
S.~Gebhardt, J.~Schellberg, R.~Lock, and W.~K{\"u}hbauch, ``Identification of
  broad-leaved dock (rumex obtusifolius l.) on grassland by means of digital
  image processing,'' \emph{Precision Agriculture}, vol.~7, no.~3, pp.
  165--178, 2006.

\bibitem{LAM21}
O.~H.~Y. Lam, M.~Dogotari, M.~Pr{\"u}m, H.~N. Vithlani, C.~Roers, B.~Melville,
  F.~Zimmer, and R.~Becker, ``An open source workflow for weed mapping in
  native grassland using unmanned aerial vehicle: Using rumex obtusifolius as a
  case study,'' \emph{European Journal of Remote Sensing}, vol.~54, no. sup1,
  pp. 71--88, 2021.

\bibitem{SIM14}
K.~Simonyan and A.~Zisserman, ``Very deep convolutional networks for
  large-scale image recognition,'' \emph{arXiv preprint arXiv:1409.1556}, 2014.

\bibitem{EVE09}
F.~Van~Evert, G.~Polder, G.~Van Der~Heijden, C.~Kempenaar, and L.~Lotz,
  ``Real-time vision-based detection of rumex obtusifolius in grassland,''
  \emph{Weed Research}, vol.~49, no.~2, pp. 164--174, 2009.

\bibitem{EVE11}
F.~K. van Evert, J.~Samsom, G.~Polder, M.~Vijn, H.-J.~v. Dooren, A.~Lamaker,
  G.~W. van~der Heijden, C.~Kempenaar, T.~van~der Zalm, and L.~A. Lotz, ``A
  robot to detect and control broad-leaved dock (rumex obtusifolius l.) in
  grassland,'' \emph{Journal of Field Robotics}, vol.~28, no.~2, pp. 264--277,
  2011.

\bibitem{BLA02}
J.~Blasco, N.~Aleixos, J.~Roger, G.~Rabatel, and E.~Molt{\'o},
  ``Ae—automation and emerging technologies: Robotic weed control using
  machine vision,'' \emph{Biosystems Engineering}, vol.~83, no.~2, pp.
  149--157, 2002.

\bibitem{WU20}
X.~Wu, S.~Aravecchia, P.~Lottes, C.~Stachniss, and C.~Pradalier, ``Robotic weed
  control using automated weed and crop classification,'' \emph{Journal of
  Field Robotics}, vol.~37, no.~2, pp. 322--340, 2020.

\bibitem{UND15}
J.~P. Underwood, M.~Calleija, Z.~Taylor, C.~Hung, J.~Nieto, R.~Fitch, and
  S.~Sukkarieh, ``Real-time target detection and steerable spray for vegetable
  crops,'' in \emph{Proceedings of the International Conference on Robotics and
  Automation: Robotics in Agriculture Workshop, Seattle, WA, USA}, 2015, pp.
  26--30.

\bibitem{BAW17}
O.~Bawden, J.~Kulk, R.~Russell, C.~McCool, A.~English, F.~Dayoub, C.~Lehnert,
  and T.~Perez, ``Robot for weed species plant-specific management,''
  \emph{Journal of Field Robotics}, vol.~34, no.~6, pp. 1179--1199, 2017.

\bibitem{FEN20}
\BIBentryALTinterwordspacing
\relax Fendt~GmbH. Neuste {G}eneration von {S}\"aroboter: {D}er {F}endt {X}aver
  wird erwachsen. [Online]. Available:
  \url{https://www.fendt.com/at/2-fendt-xaver}
\BIBentrySTDinterwordspacing

\bibitem{ECO22}
\BIBentryALTinterwordspacing
\relax ecoRobotix. {AVO} - {O}ur vision for the future: autonomous weeding (in
  development). [Online]. Available: \url{https://ecorobotix.com/en/avo/}
\BIBentrySTDinterwordspacing

\bibitem{TEN22}
\BIBentryALTinterwordspacing
\relax Tensorfield~Agriculture. Precision thermal weeding. [Online]. Available:
  \url{https://tensorfield.ag/}
\BIBentrySTDinterwordspacing

\bibitem{TAN21}
Y.~Tang, S.~Dananjayan, C.~Hou, Q.~Guo, S.~Luo, and Y.~He, ``A survey on the 5g
  network and its impact on agriculture: Challenges and opportunities,''
  \emph{Computers and Electronics in Agriculture}, vol. 180, p. 105895, 2021.

\bibitem{RAZ19}
M.~Razaak, H.~Kerdegari, E.~Davies, R.~Abozariba, M.~Broadbent, K.~Mason,
  V.~Argyriou, and P.~Remagnino, ``An integrated precision farming application
  based on 5g, uav and deep learning technologies,'' in \emph{International
  Conference on Computer Analysis of Images and Patterns}.\hskip 1em plus 0.5em
  minus 0.4em\relax Springer, 2019, pp. 109--119.

\bibitem{VAL19}
G.~Valecce, S.~Strazzella, and L.~A. Grieco, ``On the interplay between 5g,
  mobile edge computing and robotics in smart agriculture scenarios,'' in
  \emph{International Conference on Ad-Hoc Networks and Wireless}.\hskip 1em
  plus 0.5em minus 0.4em\relax Springer, 2019, pp. 549--559.

\bibitem{SMI20}
S.~Smit, T.~Tacke, S.~Lund, J.~Manyika, and L.~Thiel, ``The future of work in
  europe,'' McKinsey Global Institute, Tech. Rep., 2020.

\bibitem{RAU21}
T.~Raunholt, I.~Rodriguez, P.~Mogensen, and M.~Larsen, ``Towards a 5g mobile
  edge cloud planner for autonomous mobile robots,'' in \emph{2021 IEEE 94th
  Vehicular Technology Conference (VTC2021-Fall)}.\hskip 1em plus 0.5em minus
  0.4em\relax IEEE, 2021, pp. 01--05.

\bibitem{ANG20}
H.~Angerer, M.~Bahr, A.~Bergmann, K.~Drachsler, M.~Fessler, M.~Gergeleit,
  M.~Janker, H.~Klaus, J.~Koppenborg, C.~Schellenberger \emph{et~al.},
  ``Guidelines for 5{G} campus networks--orientation for small and medium-sized
  businesses,'' 2020.

\bibitem{OOK21}
\BIBentryALTinterwordspacing
Ookla. Germany's mobile and fixed broadband internet speeds. [Online].
  Available:
  \url{https://www.speedtest.net/global-index/germany?mobile#market-analysis}
\BIBentrySTDinterwordspacing

\bibitem{KOU19}
A.~Koub{\^a}a, A.~Allouch, M.~Alajlan, Y.~Javed, A.~Belghith, and M.~Khalgui,
  ``Micro air vehicle link (mavlink) in a nutshell: A survey,'' \emph{IEEE
  Access}, vol.~7, pp. 87\,658--87\,680, 2019.

\bibitem{EU19}
E.~Union, ``Commission implementing regulation (eu) 2019/947 of 24 may 2019 on
  the rules and procedures for the operation of unmanned aircraft, article 4
  d),'' \emph{Official Journal of the European Union}, 2019.

\bibitem{ZHA21}
J.~Zhang, J.~Maleski, D.~Jespersen, F.~Waltz~Jr, G.~Rains, and B.~Schwartz,
  ``Unmanned aerial system-based weed mapping in sod production using a
  convolutional neural network,'' \emph{Frontiers in plant science}, p. 2635,
  2021.

\bibitem{CER21}
J.~del Cerro, C.~Cruz~Ulloa, A.~Barrientos, and J.~de~Le{\'o}n~Rivas,
  ``Unmanned aerial vehicles in agriculture: A survey,'' \emph{Agronomy},
  vol.~11, no.~2, p. 203, 2021.

\bibitem{ROS2}
\BIBentryALTinterwordspacing
\relax Open~Robotics. Ros 2 documentation: Humble. [Online]. Available:
  \url{https://docs.ros.org/en/humble/}
\BIBentrySTDinterwordspacing

\bibitem{LIN73}
S.~Lin and B.~W. Kernighan, ``An effective heuristic algorithm for the
  traveling-salesman problem,'' \emph{Operations research}, vol.~21, no.~2, pp.
  498--516, 1973.

\bibitem{MAC20}
\BIBentryALTinterwordspacing
S.~Macenski, F.~Martín, R.~White, and J.~Ginés~Clavero, ``The marathon 2: A
  navigation system,'' in \emph{2020 IEEE/RSJ International Conference on
  Intelligent Robots and Systems (IROS)}, 2020. [Online]. Available:
  \url{https://github.com/ros-planning/navigation2}
\BIBentrySTDinterwordspacing

\bibitem{POE01}
E.~M. P{\"o}tsch, ``Wissenswertes zur mechanischen und chemischen
  ampferbek{\"a}mpfung.''

\bibitem{HE16}
K.~He, X.~Zhang, S.~Ren, and J.~Sun, ``Deep residual learning for image
  recognition,'' in \emph{Proceedings of the IEEE conference on computer vision
  and pattern recognition}, 2016, pp. 770--778.

\bibitem{LIN14}
T.-Y. Lin, M.~Maire, S.~Belongie, J.~Hays, P.~Perona, D.~Ramanan,
  P.~Doll{\'a}r, and C.~L. Zitnick, ``Microsoft coco: Common objects in
  context,'' in \emph{European conference on computer vision}.\hskip 1em plus
  0.5em minus 0.4em\relax Springer, 2014, pp. 740--755.

\bibitem{RED18}
J.~Redmon and A.~Farhadi, ``Yolov3: An incremental improvement,'' \emph{arXiv
  preprint arXiv:1804.02767}, 2018.

\end{thebibliography}

\end{document}